\documentclass[journal]{IEEEtran} 
\usepackage{amsmath,amssymb,amsfonts}
\usepackage{algorithmic}
\usepackage{graphicx}
\usepackage[linesnumbered,ruled,noline]{algorithm2e}
\usepackage[hidelinks]{hyperref}
\usepackage{textcomp}
\usepackage{multirow}
\usepackage{booktabs}
\usepackage{tabularx} 
\usepackage{cite}

\usepackage[protrusion=false,expansion=true]{microtype} 
\usepackage[caption=false,font=normalsize,labelfont=sf,textfont=sf,labelformat=empty]{subfig}
\usepackage[thinlines]{easytable}
\usepackage{color}
\usepackage{pifont} 
\newcommand{\cmark}{\ding{51}} 
\newcommand{\xmark}{\ding{55}} 

\begin{document}

\title{Calibration-Free EEG-based Driver Drowsiness Detection with Online Test-Time Adaptation}

\author{Geun-Deok Jang*, Dong-Kyun Han*, Seo-Hyeon Park, and Seong-Whan Lee, \IEEEmembership{Fellow, IEEE}
\thanks{This work was partly supported by Institute of Information and Communications Technology Planning and Evaluation (IITP) grants funded by the Korea government (No. RS-2019-II190079, Artificial Intelligence Graduate School Program (Korea University); No. 2021-0-02068, Artificial Intelligence Innovation Hub). (Geun-Deok Jang and Dong-Kyun Han contributed equally to this work.) (Corresponding author: Seong-Whan Lee)}
\thanks{Geun-Deok Jang and Seong-Whan Lee are with the Department of Artificial Intelligence, Korea University, Seongbuk-ku, Seoul 02841, Republic of Korea (e-mail: gd\_jang@korea.ac.kr; sw.lee@korea.ac.kr).}
\thanks{Dong-Kyun Han and Seo-Hyeon Park are with the Department of Brain and Cognitive Engineering, Korea University, Seongbuk-ku, Seoul 02841, Republic of Korea (e-mail: dk\_han@korea.ac.kr; tjgus9190@korea.ac.kr).}
\thanks{* equal contributions}
}

\markboth{IEEE Transactions on Human-Machine Systems}%
{G.-D. Jang \MakeLowercase{\textit{et al.}}: Calibration-Free EEG-based Driver Drowsiness Detection with Online Test-Time Adaptation}


\maketitle

\begin{abstract}
Drowsy driving is a growing cause of traffic accidents, prompting recent exploration of electroencephalography (EEG)-based drowsiness detection systems.
However, the inherent variability of EEG signals due to psychological and physical factors necessitates a cumbersome calibration process.
In particular, the inter-subject variability of EEG signals leads to a domain shift problem, which makes it challenging to generalize drowsiness detection models to unseen target subjects.
To address these issues, we propose a novel driver drowsiness detection framework that leverages online test-time adaptation (TTA) methods to dynamically adjust to target subject distributions. 
Our proposed method updates the learnable parameters in batch normalization (BN) layers, while preserving pretrained normalization statistics, resulting in a modified configuration that ensures effective adaptation during test time.
We incorporate a memory bank that dynamically manages streaming EEG segments, selecting samples based on their reliability determined by negative energy scores and persistence time.
In addition, we introduce prototype learning to ensure robust predictions against distribution shifts over time.
We validated our method on the sustained-attention driving dataset collected in a simulated environment, where drowsiness was estimated from delayed reaction times during monotonous lane-keeping tasks.
Our experiments show that our method outperforms all baselines, achieving an average F1-score of 81.73\%, an improvement of 11.73\% over the best TTA baseline.
This demonstrates that our proposed method significantly enhances the adaptability of EEG-based drowsiness detection systems in non-i.i.d. scenarios.

\end{abstract}

\begin{IEEEkeywords}
Driver drowsiness detection, electroencephalography, domain shift, and test-time adaptation.
\end{IEEEkeywords}

\section{Introduction}
\label{sec:introduction}
\IEEEPARstart{D}{river} drowsiness is a major concern in road safety, substantially increasing the risk of traffic accidents worldwide \cite{li2017combined}.
Most road accidents can be attributed to driver fatigue, drowsiness, and impaired judgment \cite{monteiro2019using}.
In this context, we specifically focus on driver drowsiness, the transitional state between wakefulness and sleep, distinguishing it from fatigue \cite{giorgi2024driving, ronca2022validation}.
To effectively monitor these cognitive states, electroencephalography (EEG)-based driver drowsiness detection systems have been actively studied.
EEG is a non-invasive method for recording electrical activity in the brain and directly assessing human cognitive states, making it increasingly favored for enhancing the performance of driver monitoring and assistance systems \cite{swlee_camera}.

However, a major obstacle in developing EEG-based monitoring systems is the inherent variability of EEG signals between individuals and even within the same individual across sessions. Since this variability in EEG signals complicates the construction of robust monitoring systems, a cumbersome calibration session is required \cite{d_zero_calibration}. To exclude time-consuming and labor-intensive calibration sessions, recent studies \cite{interpretable, resnet_smc} have investigated subject-independent brain-computer interface (BCI) systems, which utilize transfer learning methods such as domain adaptation (DA) and domain generalization (DG) approaches.

DA aims to minimize the domain shifts, typically using the labeled data from the source domain and unlabeled data from the target domain. However, it is infeasible to retrieve all the target subject data in monitoring systems. On the other hand, DG aims to build robust networks without any data from target distributions, which is more practical for real-world applications. DG methods mitigate domain shifts using techniques such as data augmentation \cite{resnet_smc} and domain-invariant representations \cite{d_zero_calibration}. However, DG methods make it difficult to generalize the model without prior knowledge of the unseen target domain, which often leads to limited performance.

\begin{figure}[t]
\includegraphics[width=0.5\textwidth,keepaspectratio]{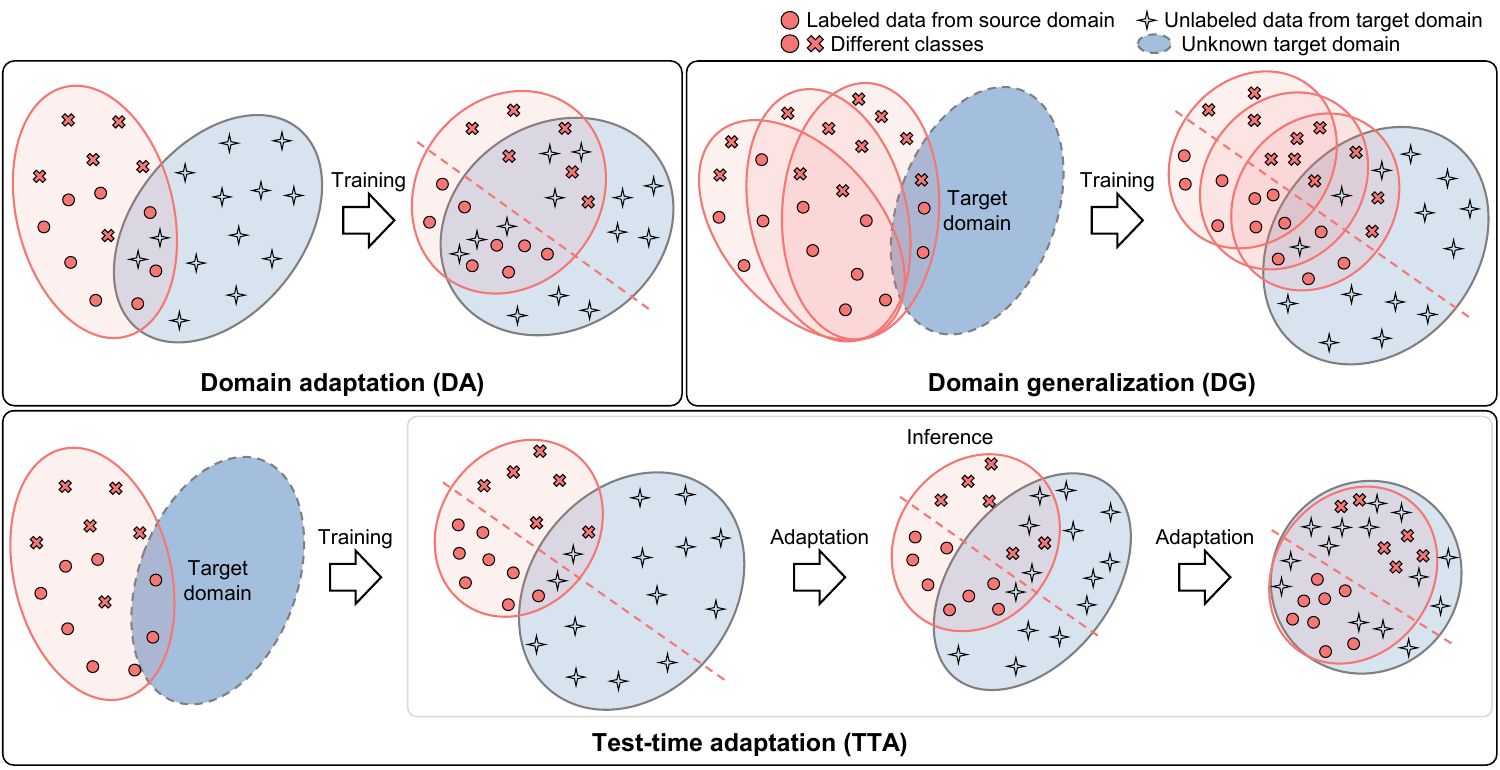}
\centering
\caption{Comparison of the transfer learning algorithms. DA utilizes both labeled source and unlabeled target domain data during training. Subsequently, DA makes a prediction with the pretrained model. DG uses only labeled source domain data during training and predicts with a pretrained model on unseen domains. TTA learns labeled source domain data during training, then dynamically adapts the pretrained model with input target domain data. This encourages effective predictions during inference with the adjusted model.}
\label{fig:compare}
\end{figure}

To overcome the limitations of domain transfer learning, several studies \cite{tent, cotta, rotta, note, sar} in the field of computer vision have introduced the test-time adaptation (TTA) algorithm, which dynamically adapts the model parameters to the target domain during inference. 
The differences among DA, DG, and TTA are illustrated in Fig.~\ref{fig:compare}. 
Inspired by these advancements, recent research has explored TTA methods in EEG-based systems such as motor imagery \cite{tta_MI} and seizure prediction \cite{TTA_seizure}.
These systems demonstrated the potential of TTA techniques to improve adaptability and performance in their respective applications. Despite these efforts, research on EEG-based driver drowsiness detection systems remains limited. Moreover, no study has yet explored online TTA methods to effectively address the challenges of an EEG-based driver drowsiness detection system.

In this study, we propose a novel driver drowsiness detection framework incorporating online TTA methods that handle the variability of EEG signals. The proposed method introduces a memory bank to store streaming target data over time, which is managed by a novel removal criterion that discards unreliable samples determined by negative energy scores and persistence time. When adapting the model, we fix the pretrained normalization statistics of the batch normalization (BN) layers and update only their affine parameters. In addition, our method incorporates prototype learning (PL) based on the memory bank data to ensure robust predictions against distribution shifts. Through our calibration-free framework, we address the critical challenge of developing an EEG-based driver drowsiness detection framework for detecting driver drowsiness. A preliminary version of this method was reported in \cite{TEMPLE}.

The primary contributions of this study are as follows:
\begin{itemize}
\item We propose a novel driver drowsiness detection framework utilizing online TTA methods, which dynamically adjusts to varying distributions and eliminates the need for labor-intensive calibration sessions.
\item Our method utilizes only the pretrained parameters from source subjects without directly accessing their raw data, ensuring the privacy of source subjects.
\item We update only the learnable parameters of the BN layers through a backward pass during model updates, thereby reducing adaptation costs.
\item Extensive experiments demonstrate the adaptability of our method to non-stationary EEG signals, highlighting its effectiveness for real-world driver drowsiness detection.
\end{itemize}

\section{Related Work}
\subsection{Subject-Independent Driver Drowsiness Detection}
EEG-based driver drowsiness detection can significantly enhance driving safety. However, the individual variability among drivers presents a considerable challenge when developing a robust driver drowsiness detection framework.
Driver drowsiness detection in the cross-subject scenario often fails to account for diverse EEG patterns exhibited by different individuals. Therefore, adaptive approaches are necessary for addressing these variations and improving the reliability of driver drowsiness detection frameworks.

Cui \textit{et al.} \cite{interpretable} presented a convolutional neural network (CNN) with separable convolutions that capture patterns of EEG signals in a spatio-temporal sequence. Paulo \textit{et al.} \cite{d_zero_calibration} demonstrated cross-subject drowsiness detection using spatio-temporal image encoding of EEG signals via recurrence plots and Gramian angular fields without requiring subject-specific calibration. Zhuang \textit{et al.} \cite{cagnn} proposed a GNN-based model that captures EEG connectivity using self-attention, enhanced by a squeeze-and-excitation block to emphasize informative frequency bands. Gong \textit{et al.} \cite{tfac_net} proposed a time-frequency attention model for single-channel EEG that highlights mental-state-relevant regions and integrates them via adaptive feature fusion.

To eliminate the need for any calibration data from the target subject, some researchers have investigated domain transfer learning methods that mitigate domain shifts.
Kim \textit{et al.} \cite{resnet_smc} augmented the data at the instance level and aligned class-related features by minimizing the distance between intra-class soft labels. 

Ding \textit{et al.} \cite{lggnet} proposed LGGNet, a neurologically inspired GNN that models both local and global brain connectivity using functional-area-based EEG graphs, combining multiscale temporal convolutions with graph filtering.

Although recent studies often use the terms ``fatigue'' and ``drowsiness'' interchangeably, following prior works that adopt the term ``drowsiness'' \cite{d_zero_calibration, cui2021subject, cagnn, resnet_smc, tfac_net, interpretable}, we consistently use the term ``drowsiness'' throughout this paper.

\subsection{Test-Time Adaptation (TTA)}\label{rel:tta}
To mitigate the existing limitations of DA and DG in the field of computer vision, several studies have optimized the model parameters for the target domain data by dynamically adapting them to unlabeled target samples during inference. This approach enhanced the adaptability and reliability in real-world applications, particularly in environments where labeled target data are scarce or unavailable.

Several studies have explored practical TTA setups applicable to real-world scenarios \cite{cotta, rotta, note, sar}. For instance, Wang \textit{et al.} \cite{cotta} introduced weight-averaged and augmentation-averaged predictions to mitigate error accumulation and catastrophic forgetting in continually changing domains. Gong \textit{et al.} \cite{note} proposed an instance-aware BN layer and prediction-balanced reservoir sampling to overcome the challenges of non-i.i.d. target data streams. They address the challenges of non-i.i.d. target data streams in online settings. Yuan \textit{et al.} \cite{rotta} proposed a robust BN layer and category-balanced memory buffer. Their removal criterion considered timeliness and uncertainty to handle practical TTA scenarios in which the distribution continually changes and correlated sampling occurs over time.

Niu \textit{et al.} \cite{sar} explored why TTA often fails in real-world scenarios, identifying that the normalization statistics in BN layers become unreliable under mixed domain shifts, small batch sizes, and imbalanced label distributions. To address this instability, they proposed a sharpness-aware and reliable entropy minimization method. This approach mitigates the impact of noisy test samples with large gradient norms, which can destabilize model adaptation.

In addition, Iwasawa \textit{et al.} \cite{t3a} have explored the use of PL to capture the underlying data structure. They demonstrated that PL is appropriate for online TTA setups, where the model needs to be adjusted for input target instances.

\begin{figure*}[t]
\includegraphics[width=0.85\textwidth,keepaspectratio]{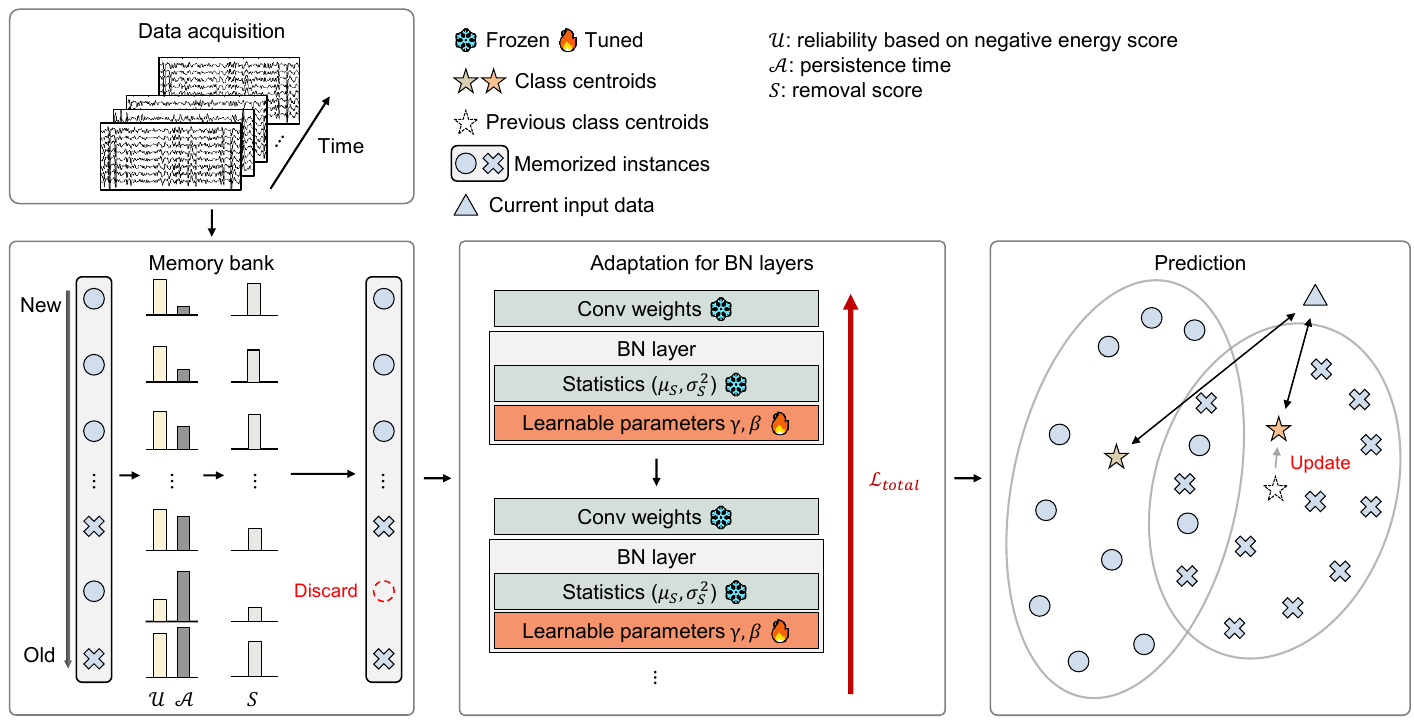}
\centering
\caption{Overview of the proposed test-time adaptation framework. Incoming EEG segments from the target subject are filtered by entropy and stored in a memory bank based on energy score and persistence. The memory bank is used to update only the learnable parameters of the BN layers via entropy minimization with energy-based regularization. Class prototypes are continuously updated from the memory to support robust prediction under distribution shifts.}
\label{fig:overview}
\end{figure*}

\subsection{TTA for EEG-based Brain-Computer Interface Systems}
In the field of EEG-based BCI systems, only a few studies have focused on online TTA to adapt BCI systems to the EEG signals from target subjects.
Mao \textit{et al.} \cite{TTA_seizure} introduced a teacher-student online TTA approach based on knowledge distillation and exponential moving average strategies. This protected the patient privacy of source subjects and allowed real-time seizure prediction for timely medical intervention.
Wimpff \textit{et al.} \cite{tta_MI} introduce an online TTA method for EEG-based motor imagery decoding to address distribution shifts between sessions and subjects without calibration data. Their approach integrates covariance alignment to align EEG covariance matrices between sessions, BN layers to replace source statistics with target statistics for better alignment with the target domain, and entropy minimization to increase prediction confidence. This combination achieves a privacy-preserving, calibration-free adaptation that effectively handles diverse EEG conditions, making BCI systems more applicable to real-world use.

Despite these efforts, research on EEG-based BCI systems that handle the distribution shifts remains limited. Moreover, no study has yet explored online TTA methods to effectively address the challenges in developing an EEG-based driver drowsiness detection system.

Among the reviewed methods, we specifically select DeepConvNet \cite{deepconvnet}, EEGNet8,2 \cite{eegnet}, ResNet1D-8 \cite{resnet_smc}, InterpretableCNN \cite{interpretable}, EEG-Conformer \cite{conformer}, CAGNN \cite{cagnn}, and LGGNet-F \cite{lggnet} as subject-independent EEG-based benchmarks, and Tent \cite{tent}, CoTTA \cite{cotta}, NOTE \cite{note}, RoTTA \cite{rotta}, T3A \cite{t3a}, SAR \cite{sar}, and Wimpff \textit{et al.} \cite{tta_MI} as representative TTA baselines. These methods are compared against our proposed approach in Section IV. EXPERIMENTS.

\section{Method}
EEG signals exhibit temporal dynamics characterized by strong temporal correlations and gradual, yet continuous, shifts in cognitive states. Unlike static or independently sampled data, EEG signals during driving reflect correlated mental states and patterns across consecutive samples. This temporal dependency violates the common assumption of sample independence inherent in many existing TTA methods, making them ineffective for EEG adaptation.
To address the non-stationarity and domain shift of driver drowsiness detection systems, we introduce a novel driver drowsiness detection framework that utilizes online TTA methods. We updated only the learnable parameters of the BN layers with entropy minimization regularized by energy-bounded loss, while preserving the pretrained normalization statistics. Our proposed framework manages a memory bank based on the confidence score, which considers the energy score and persistence time. The prototypes are then updated and constructed with the features and pseudo-labels using the memory data and the adjusted model. Predictions are made by computing the similarity between the features of the current input data and the prototypes.
In particular, the proposed framework is designed to accommodate the temporal dynamics of EEG signals, which often exhibit gradual shifts in cognitive states. By incorporating persistence-aware memory filtering and sequential prototype updates, our method continuously adapts to the evolving distribution of streaming EEG data.

Fig.~\ref{fig:overview} illustrates the overall process of our proposed framework. We describe the procedure as a pseudo-code in Algorithm \ref{algorithm}.

\subsection{Preliminaries} 
Let \( P_\mathcal{S} \) and \( P_\mathcal{T} \) denote the probability distributions of data from the source and target domains, respectively. Most domain transfer learning methods aim to solve the covariate shift between the source and target domains \( P_\mathcal{S}(\mathbf{x}) \neq P_\mathcal{T}(\mathbf{x}) \) and \( P_\mathcal{S}(y|\mathbf{x}) = P_\mathcal{T}(y|\mathbf{x}) \) \cite{tent, cotta}. However, the driver's states in the driver drowsiness detection system cause the correlation between test samples, which leads to a difference between the observed distribution over time \( P_\mathcal{T}(\mathbf{x}, y|t_1) \neq P_\mathcal{T}(\mathbf{x}, y|t_2)\) \cite{note, rotta}, making it challenging to work well on the previous TTA methods.

\subsection{Configuration of the BN layers for non-i.i.d. EEG samples}
The BN layer \cite{batchnorm} is widely used in neural network training due to its ability to promote convergence and stability. Given the input feature \(\mathbf{z} \in \mathbb{R}^{N \times C \times H \times W}\) and channel-wise statistics\footnote{We use the terms \textit{statistics} and \textit{normalization statistics} interchangeably in the context of the BN layers.} (\( \mu_c\) and \(\sigma^2_c\)), BN layers normalize the feature for each channel:

\begin{equation}
    \text{BN}(\mathbf{z_{:, c, :, :}}; \mu_c, \sigma^{2}_c) = \gamma_C \frac{\mathbf{z_{:, c, :, :}} - \mu_c}{\sqrt{\sigma^{2}_c +\epsilon}} + \beta_c,
\end{equation}
where \(\gamma_c\) and \(\beta_c\) denote channel-wise learnable parameters (scale and shift) and $\epsilon$ denotes a small positive constant that prevents numerical instability.

Given the channel-wise mini-batch statistics (\(\hat{\mu}_c\) and \(\hat{\sigma}_c^{2}\)) calculated from the input data, the normalization statistics of the BN layers are continuously updated using a momentum term:

\begin{subequations}
\begin{align}
    \hat{\mu}_c &= \frac{1}{NHW} \sum_{n, h, w} \mathbf{z}_{n, c, h, w}, \label{eqn:batch_mean} \\
    \hat{\sigma}_c^{2} &= \frac{1}{NHW} \sum_{n, h, w} (\mathbf{z}_{n, c, h, w} - \hat{\mu}_c)^2, \label{eqn:batch_var} \\
    \mu_c &= \delta \cdot \mu_c + (1 - \delta) \cdot \hat{\mu}_c, \label{eqn:update_mean} \\
    \sigma_c^{2} &= \delta \cdot \sigma_c^{2} + (1 - \delta) \cdot \hat{\sigma}_c^{2}, \label{eqn:update_var} 
\end{align}
\end{subequations}
where \(\delta\) is the momentum term that balances the influence of new mini-batch statistics against the running estimates.

Following recent studies \cite{tent, rotta, note, cotta, tta_MI}, we focus on optimizing the BN layers on a given feature extractor during inference. Existing TTA methods can be categorized based on how they apply normalization statistics in BN layers.
Some studies \cite{tent, cotta, tta_MI} normalize target data using only mini-batch statistics (\ref{eqn:batch_mean} and \ref{eqn:batch_var}) for both updating the parameters and prediction.
Conversely, other studies \cite{rotta, note} use a continuous tracking of statistics from input target features to update parameters (\ref{eqn:update_mean} and \ref{eqn:update_var}). Subsequently, they apply these updated statistics for prediction.

While both approaches use channel-wise statistics to update learnable parameters based on target domain data, they differ in how they handle statistics for prediction: the first approach relies on the current batch statistics, whereas the second continuously updates statistics at each time step.
However, estimating statistics from individual streaming instances (batch size of 1) can ruin BN statistics \cite{sar}. In addition, estimating statistics from correlated samples may introduce a bias toward classes that dominate the distribution \cite{rotta, note}.

To address this issue, we update only the learnable parameters within the BN layers using input test samples, while preserving the pretrained statistics from the multiple source subjects. This approach marks a key distinction from prior TTA studies, which typically adjust both parameters and statistics during adaptation. Given the feature, our proposed BN layers compute:

\begin{equation}
    \text{BN}_{\text{ours}}(\mathbf{z_{:, c, :, :}}; \mu_c^{\mathcal{S}}, (\sigma_c^{\mathcal{S}})^2) = \gamma_c \frac{\mathbf{z_{:, c, :, :}} - \mu_c^{\mathcal{S}}}{\sqrt{(\sigma_c^{\mathcal{S}})^2 +\epsilon}} + \beta_c,
\end{equation}
where \(\mu_c^{\mathcal{S}}\) and \((\sigma_c^{\mathcal{S}})^2\) indicate source pretrained statistics, which are fixed during the adaptation process.

\subsection{Entropy Minimization with Energy-bounded Loss}\label{method:ment}

We optimize the BN parameters using the entropy minimization, which is widely used in TTA to adapt to the target distribution \cite{tent, note, rotta}:

\begin{equation}
    \mathcal{L}_{ent} = -\frac{1}{\mathcal{N}} \sum_{i=1}^{\mathcal{N}} \sum_{k=1}^{\mathcal{C}} p_{i,k} \log p_{i,k},
\label{eqn:entropy_loss}
\end{equation}
where \(p_{i,k}\) denotes the probability that an instance belongs to class \(k\).

In addition to the entropy minimization, we introduce the energy-bounded loss as a regularization term:

\begin{equation}
    E(\mathbf{x}; f; \tau) = - \log \sum_{k=1}^{\mathcal{C}} e^{f_{k}(\mathbf{x})/(\tau^2)}, \label{eqn:energy}
\end{equation}
\begin{align}
    \mathcal{L}_{energy} &= \mathop{\mathbb{E}} \left[\max(0, E(\mathbf{x}; f) - m_{in})\right]^2 \notag \\
    &+ \mathop{\mathbb{E}} \left[\max(0, m_{out} - E(\mathbf{x}'; f))\right]^2, \label{eqn:energy_loss}
\end{align}
where \( \mathbf{x} \) represents the input data, \(\mathbf{x}'\) denotes the augmented data used to simulate domain shifts, \(E(.)\) is the energy function, \(\tau\) is the temperature scaling factor (set to 1 in energy-bounded loss), \(m_{in}\) is the margin for in-distribution samples, \(m_{out}\) is the margin for out-of-distribution samples, \(f\) represents the neural network, and \(\mathcal{C}\) is the total number of classes.

The energy-bounded loss encourages the model to treat the target samples as in-distribution by assigning lower energy scores to them \cite{energyOOD2020}. To explicitly model potential distribution shifts, we apply augmentations to the test samples. Specifically, the loss penalizes original target samples whose energy scores exceed the threshold \(m_{in}\) and augmented (shifted) samples whose scores fall below the threshold \(m_{out}\).

The parameters are optimized using the following comprehensive objective function:
\begin{equation}
    \mathcal{L}_{total} = \lambda_{ent} \mathcal{L}_{ent} + \lambda_{eng} \mathcal{L}_{energy} ,
\label{eqn:loss_total}
\end{equation}
where \(\lambda_{ent}\) and \(\lambda_{eng}\) denote the weights for balancing loss.

\subsection{Memory Bank}\label{method:memory}
The reliability of normalization statistics is highly dependent on the batch size of the input data \cite{sar}. Using a smaller batch size may hinder the network's ability to capture the target distribution \cite{MemoryBuffer}, posing significant challenges for practical driver drowsiness detection systems. Specifically, in an EEG-based driver drowsiness detection system, estimating and updating statistics can be affected by the individual features \(\mathbf{z} \in \mathbb{R}^{1 \times C \times H \times W}\). Furthermore, the correlation among target instances presents discrepancies between the observed distribution and the genuine target distribution, leading to inaccurate statistics of the BN layers \cite{rotta, note}.

To address this challenge, we propose a memory bank with a novel removal criterion that utilizes the energy score and persistence time. When the first input arrives, instead of waiting to fill the memory bank over time, we initialize it by applying augmentation algorithms to the input. Specifically, we apply white Gaussian noise and a permutation algorithm \cite{ijcnn_contrastive_sleep} that mixes segments of the input EEG signals to generate diverse augmented samples simulating the target distribution. 
When the memory bank is full, we compute the negative energy score \( - E(\mathbf{x})\) \cite{energyOOD2020} for each sample to identify and remove unreliable samples from the memory.
The energy score acts as a confidence measure, with lower scores indicating samples that the model considers less reliable or potentially out-of-distribution.
In addition to the negative energy score, we further regularize it with persistence time, which indicates how long the data have been stored in the memory bank.
This allows the removal mechanism to penalize not only low-confidence samples but also outdated entries that no longer reflect the current distribution.
We use the persistence time as a parameter for the temperature of the negative energy score, gradually decreasing the confidence of outdated samples over time. The overall removal score is as follows:

\begin{equation}
    S(\mathbf{x} \mid \mathbf{x} \in \mathcal{M}) = -E(\mathbf{x}; f,\mathcal{A}) = \log \sum^\mathcal{C}_{k=1} e^{f_k(\mathbf{x}) / (\mathcal{A}^2)},
\label{eqn:removal}
\end{equation}
where \(\mathbf{x}\) indicates samples in the memory bank, \(\mathcal{M}\) represents the memory bank, and \(\mathcal{A}\) represents the persistence time.

Our proposed removal criterion selectively removes overconfident and outdated samples, allowing the memory bank to align with the current distribution of the streaming data.

\begin{algorithm}[t!]
\caption{Proposed Online Test-Time Adaptation}
\label{algorithm}
\textbf{Require:} Streaming input EEG sample $\mathbf{x}_t$ at time $t$ ($t \leq T$); Memory bank $\mathcal{M}$ with capacity $\mathcal{N}$; Occupancy $\mathcal{O}$ of $\mathcal{M}$; Pretrained BN statistics $(\mu_c^{\mathcal{S}}, (\sigma_c^{\mathcal{S}})^2)$; Learnable affine parameters $(\gamma_c, \beta_c)$ in BN layers (scale and shift)

\For{$t=1$ \KwTo $T$}{
    \tcp{1. Store current input}
    Insert $\mathbf{x}_t$ into $\mathcal{M}$

    \tcp{2. Initialize memory bank}
    \If{$t = 1$}{
        \While{$\mathcal{O} < \mathcal{N}$}{
            Generate augmentations of $\mathbf{x}_t$ (Gaussian noise \& permutation)

            Add augmented samples to $\mathcal{M}$
        }
        Initialize prototypes using pretrained classifier weights

    }

    \tcp{3. Memory bank update}
    Compute removal score all samples in $\mathcal{M}$ via \eqref{eqn:removal}

    Discard the sample with the highest removal score

    \tcp{4. BN parameter adaptation}

    \textbf{Preserve} source pretrained statistics $(\mu_c^{\mathcal{S}}, (\sigma_c^{\mathcal{S}})^2)$

    Update BN affine parameters $(\gamma_c, \beta_c)$ via \eqref{eqn:loss_total}

    \tcp{5. Prototype learning}

    Compute pseudo prototypes $\hat{\mathcal{P}}_k$ for each class $k$ 

    Update each prototype $\mathcal{P}_k$ via \eqref{eqn:ema}

    \tcp{6. Prediction}
    Prediction based on the similarity as in \eqref{eqn:predict}
}
\end{algorithm}

\subsection{Prototype Learning }\label{method:pl}
Prototype learning (PL) aligns with the principles of few-shot learning, where a model learns from limited labeled data. However, unlike few-shot learning, online TTA operates with only a single unlabeled target sample available at each time step. Since PL generally involves updating class center representations, it is computationally efficient and has low latency, making it suitable for real-world applications.

In our approach, we combine the PL with a memory bank that captures the current target distributions. To initialize the prototypes, we leverage the weights from a source pretrained classifier, incorporating source knowledge into the prototypes. The prototypes are created by averaging the features and pseudo-labels of the memory data using the adapted model.

To ensure reliable prototype estimation, we consider only the memory samples that are pseudo-labeled as class \(k\) and have energy scores above the threshold \(m_{out}\). Let \(\mathcal{M}_k\) denote the set of such samples:

\begin{equation}
\mathcal{M}_k = \left\{ \mathbf{x}_i \in \mathcal{M} \mid \hat{\mathbf{y}}_i = k,\; E(\mathbf{x}_i) > m_{out} \right\},
\label{eqn:filter_samples}
\end{equation}
where \(\hat{\mathbf{y}}_i\) is the pseudo-label assigned to sample \(\mathbf{x}_i\).

The pseudo prototype \(\hat{\mathcal{P}}_k\) for class \(k\) is computed by averaging the feature representations \(\mathbf{z}_i\) of all samples in \(\mathcal{M}_k\):

\begin{equation}
\hat{\mathcal{P}}_{k} = \frac{1}{|\mathcal{M}_k|} \sum_{\mathbf{x}_i \in \mathcal{M}_k} \mathbf{z}_i.
\end{equation}

We then update the final prototype \(\mathcal{P}_k\) using exponential moving average (EMA):

\begin{equation}
\mathcal{P}_{k} = \alpha \mathcal{P}_{k} + (1 - \alpha) \hat{\mathcal{P}}_{k},
\label{eqn:ema}
\end{equation}
where \(\alpha\) is the smoothing factor that controls the update rate.

Finally, the prediction is made by measuring the similarity between the current input feature \(\mathbf{z}_t\) and each class prototype \(\mathcal{P}_k\) via dot-product attention:

\begin{equation}
\arg\max_k P(y = k \mid \mathbf{z}_t) = \frac{\exp \left( \mathbf{z}_t \cdot \mathcal{P}_k \right)}{\sum_j \exp \left( \mathbf{z}_t \cdot \mathcal{P}_j \right)}.
\label{eqn:predict}
\end{equation}

\section{Experiments}
\subsection{Materials} 
\begin{figure}[t]
\includegraphics[width=0.5\textwidth]{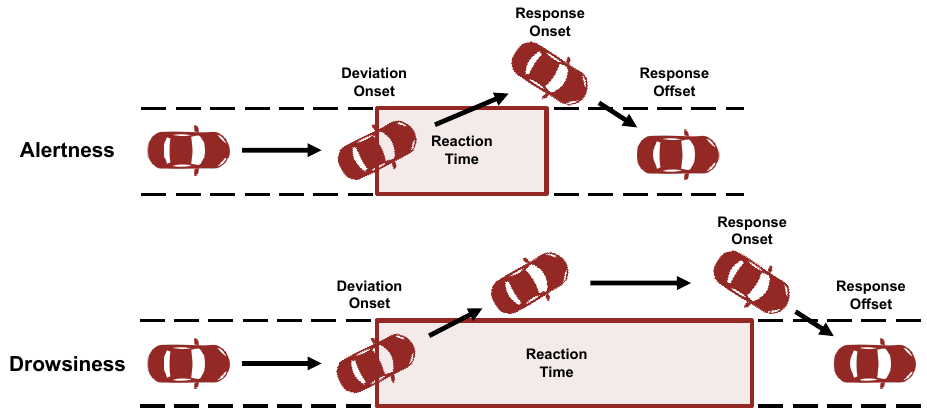}
\centering
\caption{Experimental paradigms of the sustained-attention driving dataset were conducted in the following sequence for each trial: response offset, deviation onset, and response onset. Drowsy drivers exhibited longer RTs compared to alert drivers.}
\label{fig:dataset}
\end{figure}

We used the preprocessed version of the sustained-attention driving dataset from a previous study \cite{interpretable, dataset_drowsiness}. The original dataset comprises 62 sessions collected from 27 healthy participants, each of whom engaged in a 90-minute driving task using a virtual reality (VR) driving simulator. During the experiment, participants were instructed to maintain their vehicle at the center of the lane, while the system introduced unexpected lane-departure events to simulate a non-ideal road surface that caused the car to drift left or right. Participants were required to respond immediately by steering the vehicle back to the center lane, simulating the continuous vigilance required in real driving. The durations of lane departure, steering, and return to the center of the lane are referred to as deviation onset, response onset, and response offset, respectively.
In the absence of camera-based measures (e.g., eye closure), these lane-departure events served as the primary behavioral probes to quantify drowsiness via reaction times. Furthermore, occurring at randomized intervals, these events established a monotonous task environment designed to induce passive drowsiness throughout the 90-minute session.
Each lane-departure event was treated as a distinct trial and segmented accordingly. Experimental paradigms of the sustained-attention driving dataset are illustrated in Fig.~\ref{fig:dataset}.

EEG data were continuously recorded at 500 Hz using a wired cap with 30 EEG electrodes and 2 reference electrodes, which were placed according to a modified international 10–20 system. EEG signals were preprocessed with a 1–50 Hz band-pass filter. To address the contamination of EEG signals by eye blinks, the recordings were manually inspected, and segments containing apparent blink artifacts were visually rejected. Ocular and muscular artifacts were further removed using the Automatic Artifact Removal (AAR) plugin of EEGLAB. The preprocessed EEG data were then downsampled to 128 Hz, and 3-second segments were extracted prior to each deviation onset to capture neural activity preceding behavioral responses.

To ensure data quality and class balance, additional filtering steps were performed as follows. First, sessions with less than 50 samples in either class were excluded. Second, if there were multiple sessions from the same subject, the session with the most balanced class distribution was selected. Both raw and preprocessed datasets are publicly available.

According to prior studies \cite{d_zero_calibration, d_RT_label, d_decomposition, inter_subject_fatigue, interpretable}, we followed two metrics for each sample. Local reaction time (RT) is the duration taken to respond to a lane deviation event. Global RT is the average of RTs recorded within the 90-second window before the lane departure event. Samples with both local RT and global RT shorter than 1.5 times the alert RT were classified as alertness, whereas those with both longer than 2.5 times the alert RT were classified as drowsiness. To ensure the balance and integrity of the data, sessions with fewer than 50 samples in either class were excluded. The session with the most balanced class distribution was selected if multiple sessions were available for the same participant.

\subsection{Baselines}
We established two categories of baselines for comparison. We referred to the pretrained models evaluated directly on the target subject without any adaptation methods as ``Source''. ``Source'' includes DeepConvNet \cite{deepconvnet}, EEGNet8,2 \cite{eegnet}, ResNet1D-8 \cite{resnet_smc}, InterpretableCNN \cite{interpretable}, EEG-Conformer \cite{conformer}, CAGNN \cite{cagnn}, and LGGNet-F \cite{lggnet}.

Specifically, we adopted a diverse set of ``Source'' baselines that use conventional CNN-based architectures (DeepConvNet \cite{deepconvnet}, EEGNet8,2 \cite{eegnet}, ResNet1D-8 \cite{resnet_smc}, InterpretableCNN \cite{interpretable}), graph-based architecture (CAGNN \cite{cagnn}), and advanced Transformer variants (LGGNet \cite{lggnet}, EEG-Conformer \cite{conformer}). Some of them have been previously validated on the same sustained-attention driving dataset in their respective publication. These models serve as strong baselines that reflect both foundational and state-of-the-art approaches in EEG-based BCI systems.

For ``TTA'' baselines, we selected methods that represent a broad range of TTA strategies. Specifically, BN adaptation approaches (Tent \cite{tent}, SAR \cite{sar}), memory-based and augmentation-aware methods (CoTTA \cite{cotta}, NOTE \cite{note}, RoTTA \cite{rotta}), prototype-based classifiers (T3A \cite{t3a}), and EEG-specific online TTA (Wimpff \textit{et al.} \cite{tta_MI}) were evaluated.

Each TTA method was implemented to adapt the same source-pretrained EEGNet8,2 \cite{eegnet} architecture to unseen target subjects under identical hyperparameter settings and evaluation protocols, ensuring a fair and meaningful comparison.

\begin{table*}[ht]
    \centering
    \caption{Overall F1-score (\%) comparison of driver drowsiness detection on a sustained-attention driving dataset}
    \label{tab:comparison}
    \resizebox{\textwidth}{!}{%
    \renewcommand{\arraystretch}{1.25}
        \begin{tabular}{llcccccccccccc}
        \toprule
        
        Category & Method & S1$^{*}$ & S2 & S3 & S4 & S5 & S6 & S7 & S8 & S9 & S10 & S11 & Avg. (std.) \\ \midrule
        
        \multirow{8}{*}{Source}
        & DeepConvNet \cite{deepconvnet} & 81.44 & 8.66 & 56.13 & 62.50
        & 70.71 & 81.00 & 80.79 & 62.61
        & 77.19 & 49.32 & 60.91 & 62.84 (21.06) \\
        & EEGNet8,2 \cite{eegnet} & 73.86 & 29.52 & 62.04 & 68.97
        & 76.04 & 87.04 & 83.02 & 64.23
        & 77.23 & 52.50 & 65.12 & 67.23 (15.96) \\
        & ResNet1D-8 \cite{resnet_smc} & 81.97 & 33.46 & 50.98 & 66.02
        & 52.63 & 73.91 & 80.57 & 67.12
        & 77.66 & 60.98 & 75.76 & 65.55 (15.00) \\
        & InterpretableCNN \cite{interpretable} & 67.47 & 19.33 & 64.23
        & 66.67
        & 80.00 & 73.85 & 83.81 & 65.59
        & 80.79 & 54.55 & 75.76 & 66.55 (17.89) \\ 
        & EEG-Conformer \cite{conformer} & 72.39 & 31.03 & 42.42 & 58.49
        & 73.24 & 84.47 & 77.55 & 63.32
        & 75.09 & 62.65 & 74.45 & 65.01 (16.01) \\
        & CAGNN \cite{cagnn} & 77.97 & 31.18 & 26.67 & 49.75
        & 26.47 & 68.37 & 52.90 & 48.30
        & 65.43 & 65.12 & 65.99 & 52.56 (17.98) \\
        & LGGNet-F \cite{lggnet} & 67.00 & \textbf{36.69} & 49.61 & 57.97
        & 68.42 & 68.75 & 74.87 & 63.59
        & 77.56 & 53.33 & 71.91 & 62.70 (12.25) \\
        
        \midrule
                                     
        \multirow{8}{*}{TTA} 
        & Tent \cite{tent} & 3.51 & 28.21 & 1.10 & 12.00
        & 0.00 & 7.75 & 1.83 & 0.00
        & 0.00 & 11.02 & 5.59 & 6.46 (8.43) \\
        & CoTTA \cite{cotta} & 16.82 & 27.69 & 1.10 & 19.78 
        & 4.96 & 4.69 & 0.00 & 9.94
        & 3.75 & 16.36 & 10.07 & 10.47 (8.75)  \\
        & T3A \cite{t3a} & 75.98 & 34.11 & 67.13 & \textbf{69.70} 
        & 79.40 & 87.11 & 82.63 & 70.50
        & 81.90 & 51.22 & 70.27 & 70.00 (15.39) \\ 
        & NOTE \cite{note} & 40.25 & 9.26 & 32.62 & 57.14  
        & 24.05 & 67.89 & 56.97 & 29.91
        & 1.10 & 45.21 & 33.14 & 36.14 (20.34) \\ 
        & RoTTA \cite{rotta} & 68.97 & 5.80 & 62.59 & 49.03 
        & 83.61 & 89.27 & 83.02 & 46.67
        & 20.41 & 0.00 & 65.18 & 52.23 (31.30) \\
        & SAR \cite{sar} & 39.05 & 6.11 & 56.15 & 65.85 
        & 77.25 & 88.29 & 31.01 & 66.14
        & 75.72 & 51.95 & 48.42 & 55.09 (23.56) \\
        & Wimpff \textit{et al.} \cite{tta_MI} & 44.44 & 25.36 & 42.54 & 40.00 
        & 45.19 & 54.82 & 46.84 & 40.43
        & 46.26 & 31.06 & 45.28 & 42.02 (7.98) \\ \cmidrule{2-14}
        
        & Ours
        & \textbf{94.68}
        & 28.33 
        & \textbf{95.63}  
        & 64.49
        & \textbf{95.73} 
        & \textbf{94.42} 
        & \textbf{84.51} 
        & \textbf{87.29}
        & \textbf{88.66} 
        & \textbf{75.27} 
        & \textbf{90.04} 
        & \textbf{81.73 (20.16)} \\ \bottomrule
          
        \end{tabular}
    }
    \par
    \vspace{1mm}
    \raggedright 
    \footnotesize 
    \textit{*} S: Subject
\end{table*}

\subsection{Implementation Details} \label{exp_settings}
To evaluate the robustness of our TTA methods, we established a sophisticated experimental setup that contained both pretraining and inference stages. Our experiments employed leave-one-subject-out (LOSO) cross-validation \cite{LOSO}, where one subject was designated as the target subject for inference, while the others were used for pretraining. This process was repeated for each subject.

To guide the experimental design of this study, we selected benchmark models from the reviewed literature that satisfy two key criteria. First, public availability of the source code. Second, compatibility with the subject-independent LOSO cross-validation protocol used in our EEG-based drowsiness detection task. These choices ensure reproducibility and fairness in performance comparison.

During the pretraining phase, we utilized the Adam optimizer with a learning rate of 0.001. We set the batch size to 32 and trained the model for 100 epochs. This process was performed for each target subject following a LOSO cross-validation.

An essential component of our proposed framework was the configuration of the BN layers after pretraining. As described in Section~\ref{method:ment}, we ensured that the BN layers do not alter the normalization statistics pretrained with multiple source subjects, while focusing on updating only the learnable parameters.

During the adaptation process with the streaming target data, we used the source pretrained EEGNet8,2 \cite{eegnet}, which showed the highest performance among all source baselines without using any adaptation methods. We utilized the AdamW optimizer \cite{adamw} with a learning rate of 0.001 and a weight decay of 0.1. We used a test batch size of one and set the adaptation step to one, minimizing the latency of the adaptation process. We set the margin values \(m_{in}\) and \(m_{out}\) in \eqref{eqn:energy_loss} to -15 and -7, respectively. Additionally, we set the smoothing factor \(\alpha\) to 0.9 to facilitate the update of the adaptive prototypes. Finally, we set the balance weights \(\lambda_{ent}\) and \(\lambda_{eng}\) in \eqref{eqn:loss_total} to 2 and 0.01, respectively.

For the TTA baselines, we used the hyperparameter values reported in their respective publications or those provided in their official implementation code. We set the memory bank capacity to 16 for all TTA baselines that used a memory bank, based on preliminary experiments where we tested capacities from 8 to 64 and found 16 to yield the best performance.

We conducted experiments on an NVIDIA RTX 3090 GPU.

\subsection{Results}
In this study, we assessed the effectiveness of our proposed framework against the source baselines and various TTA baselines on a sustained-attention driving dataset \cite{interpretable}. We evaluated all methods using the F1-score metric, which represents the harmonic mean of precision and recall. F1-score is particularly useful in scenarios with imbalanced label distributions. In Table~\ref{tab:comparison}, we evaluated the performances of our proposed method compared to source baselines and TTA baselines on a sustained-attention driving dataset.

The ``Source'' methods achieve an average F1-score ranging from 62.84\% to 67.23\%. Among the TTA methods, Tent \cite{tent} shows a marked performance degradation, achieving an average F1-score of 6.46\%. Due to its dependence on mini-batch statistics for individual input instances, this causes model collapse across all target subjects \cite{sar}.
CoTTA \cite{cotta} obtains an average F1-score of 10.47\%. Its performance is very low across all target subjects, indicating issues with normalization statistics when using only input instances.
T3A \cite{t3a}, designed for online TTA using PL, achieves an average F1-score of 70\%, demonstrating the potential of PL for driver drowsiness detection. However, storing all past data in T3A can lead to memory and computational constraints and may retain outdated information as the target distribution evolves.
NOTE \cite{note} shows a decrease in performance with an average F1-score of 36.14\%. This indicates that its proposed prediction-balanced memory bank failed to mitigate the label distribution shifts in the driver drowsiness detection task, due to the dominance of certain classes over time.
RoTTA \cite{rotta} performs better with an average F1-score of 52.23\%. However, it still does not surpass the performance of the source models that did not attempt to adjust to the target distribution. This performance decrease is attributed to the updating of global statistics based on the target data, which can introduce noise and reduce robustness.
Although SAR \cite{sar} was designed for online TTA by introducing batch-agnostic normalization layers that prevent a collapse of normalization statistics, SAR decreases performance with an average F1-score of 55.09\% even after the adaptation process.
The online TTA method proposed by Wimpff \textit{et al.} \cite{tta_MI} shows a decreased performance with an average F1-score of 42.02\%, indicating that using the target statistics ruins the normalization statistics of the BN layers.

Our proposed method notably outperforms all other baselines, achieving the highest average F1-score of 81.73\%. This shows an average improvement of approximately 11\% over the best performance among the TTA baselines.
This significant improvement validates the effectiveness of our approach in a strictly calibration-free setting. By accounting for the non-stationary and temporally correlated nature of EEG signals, our framework enables robust adaptation to target subjects, effectively overcoming the domain shifts that typically challenge subject-independent models.
The average per-sample inference time was measured at 15.94 ms.

\begin{table}[t]
\centering
\caption{Performance comparison of varying ablation settings}
\label{tab:ablation-study}
\renewcommand{\arraystretch}{1.25} 
\begin{tabularx}{\columnwidth}{Xcccc}
\toprule
Method & F1-score & AUROC & Precision & Recall \\ 
\midrule 
EEGNet8,2 (source) & 67.23 & 73.18 & 74.90 & 67.46 \\
Ours w/o BN updates & 69.23 & 75.66 & \textbf{80.22} & 68.77 \\
Ours w/o memory and PL & 75.06 & 77.84 & 70.92 & 86.95 \\
Ours w/o PL & 75.84 & 78.40 & 71.32 & 87.84 \\
Ours & \textbf{81.73} & \textbf{83.83} & 79.75 & \textbf{91.06} \\
\bottomrule
\end{tabularx}
\end{table}

\begin{table}[t]
\caption{Comparison of existing BN configurations}
\label{tab:configuration}
\centering
\resizebox{\columnwidth}{!}{%
\begin{tabular}{lcccccc}
\toprule
Method & \begin{tabular}[c]{@{}c@{}}Track\\ Statistics\end{tabular} & \begin{tabular}[c]{@{}c@{}}Initialize\\ Statistics\end{tabular} & F1-score       & AUROC          & Precision      & Recall         \\ \midrule

Tent \cite{tent} & \multirow{3}{*}{\xmark} & \multirow{3}{*}{\cmark} & 6.46 & 46.31 & 22.73 & 9.11 \\ 
CoTTA \cite{cotta} & & & 10.47 & 50.15 & 48.72 & 9.00 \\ 
Wimpff \textit{et al.} \cite{tta_MI} & & & 42.02 & 52.74 & 48.20 & 41.26 \\ \midrule
NOTE \cite{note} & \multirow{2}{*}{\cmark} & \multirow{2}{*}{\xmark} & 36.14 & 53.65 & 46.53 & 33.50 \\
RoTTA \cite{rotta} & & & 52.23 & 65.14 & 59.79 & 49.09 \\ \midrule

\multirow{4}{*}{Ours} & \xmark & \cmark & 0.00 & 48.69 & 0.00 & 0.00 \\ \cmidrule{2-7}
& \cmark & \xmark & 40.72 & 52.29 & 43.78 & 42.98 \\ \cmidrule{2-7}
& \xmark & \xmark & \textbf{81.73} & \textbf{83.83} & \textbf{79.75} & \textbf{91.06} \\
\bottomrule
\end{tabular}%
}
\end{table}

\subsection{Ablation Study}
We performed an ablation study to evaluate the individual contribution of each component by removing it. The ``EEGNet8,2 (source)'' configuration serves as the baseline. The performance of the ablation study for our proposed method is illustrated in Table~\ref{tab:ablation-study}.

In the scenario where we did not update the BN layers, referred to as ``Ours w/o BN updates'', we evaluated the proposed PL method based on the memory bank for TTA. This configuration demonstrates a performance improvement over the ``EEGNet8,2 (source)'' model, highlighting the effectiveness of PL for driver drowsiness detection in TTA settings.
When we update only the learnable parameters of the BN layers, termed as ``Ours w/o memory and PL'', we observe significant increases in F1-score and recall over the source baseline. This indicates that the BN layers can effectively adapt to even a single target instance of the target domain by preserving the source normalization statistics, showing their robustness against smaller batch sizes and distribution shifts.
``Ours w/o PL'' shows slight improvements across all metrics compared to ``Ours w/o memory and PL''. This indicates that introducing the memory bank to update the BN layers enhances stability, compared to updating the model based on individual instances.

Our proposed framework exhibits significant performance improvements across all metrics. Notably, the F1-score increases to 81.73\%, which highlights the strong capability of the model to discriminate between the driver's mental states under domain shift environments.

\begin{figure*}[t]
\centering
\subfloat[]{\includegraphics[width=0.9\textwidth]{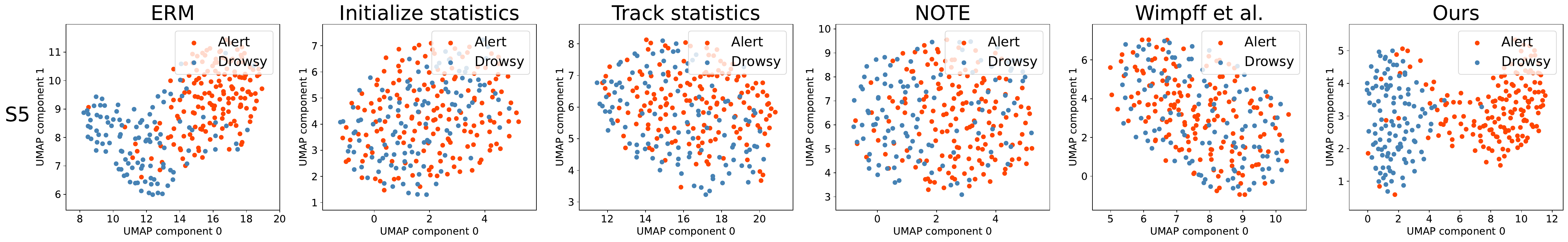}} \\[-3.5ex] 
\subfloat[]{\includegraphics[width=0.9\textwidth]{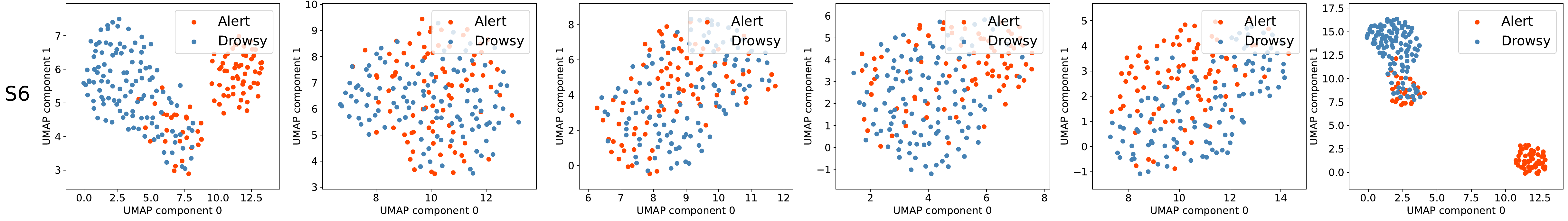}} \\[-3.5ex]
\subfloat[]{\includegraphics[width=0.9\textwidth]{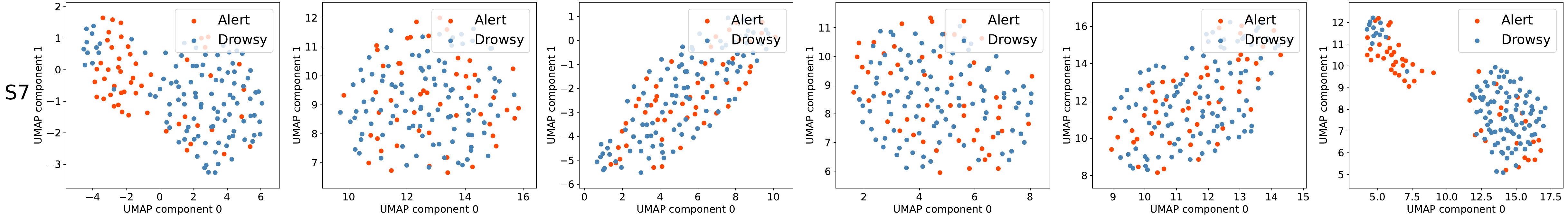}}
\caption{UMAP feature visualization from target subjects. The row and column indicate the target subjects and adaptation methods, respectively. The first column shows the features from the ERM (no adaptation) model. The second and third columns show the features when adapting the model using existing BN configurations. The fourth and fifth columns show the features when adapting the model using TTA methods. The last column shows the features of our proposed method. We show that the proposed method significantly enhances performance and disentanglement.}
\label{fig:viz}
\end{figure*}

\begin{figure}[t]
\includegraphics[width=0.5\textwidth,keepaspectratio]{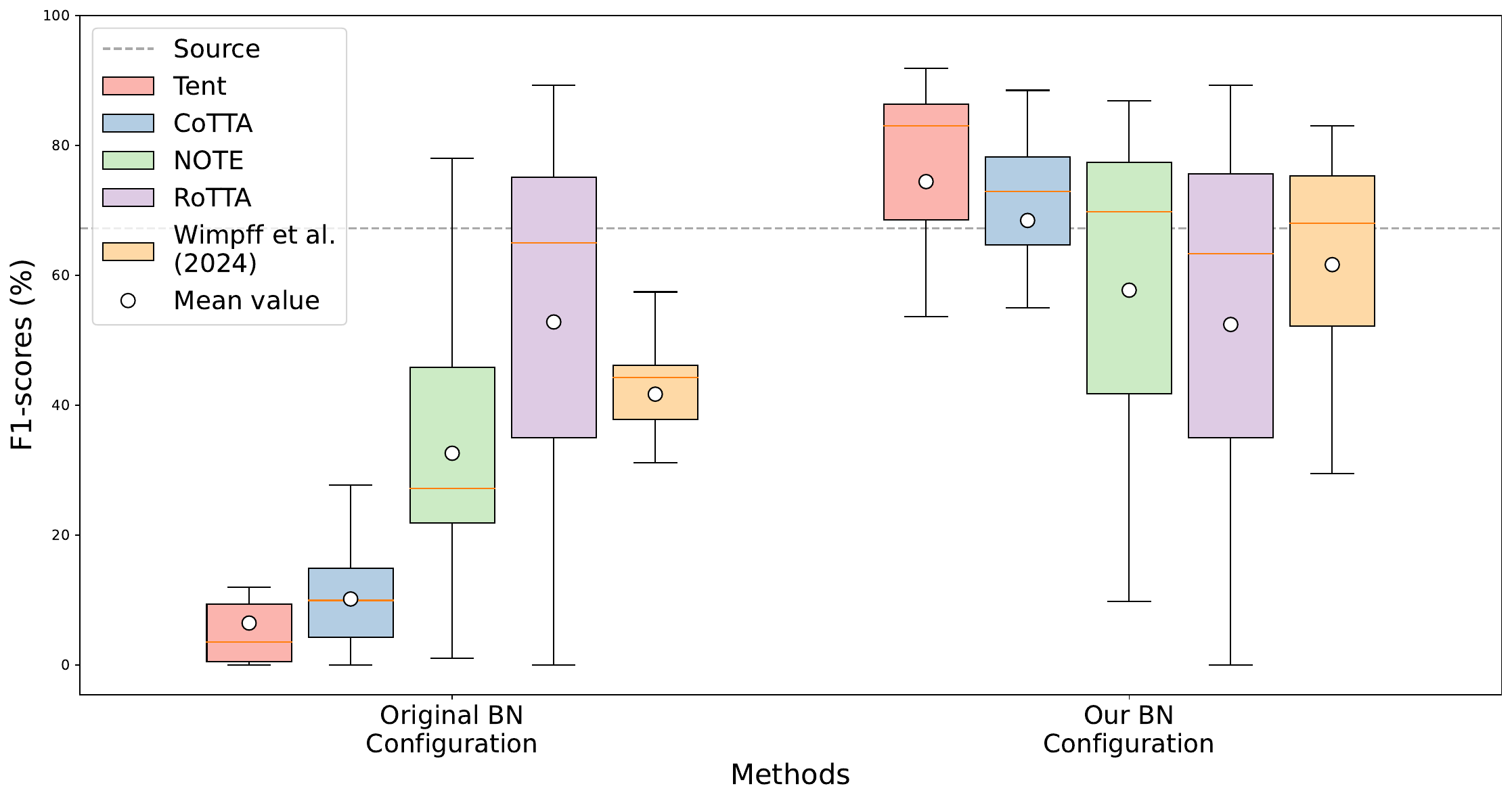}
\centering
\caption{Performance comparison when applying our BN configuration to the existing TTA baselines. The gray horizontal line denotes the average F1-score of the source pretrained EEGNet8,2 without adaptation methods. Existing baselines with their original BN configuration are presented on the left, while the baselines with our BN configuration are shown on the right. The overall performance of the existing TTA baselines significantly improves after manipulating the BN layers.}
\label{fig:reconf}
\end{figure}

\subsection{Normalization Configurations of the BN Layers}
In Table~\ref{tab:configuration}, we compared the performance of the proposed method when adapting the various normalization configurations of the BN layers that were employed in the existing TTA baselines \cite{tent, cotta, note, rotta, tta_MI}.
First, we configured the BN layers to normalize the input samples using only input batch statistics. Then, we transformed the input with updated learnable parameters for prediction. However, as illustrated in Table~\ref{tab:configuration}, this setup showed the most severe performance degradation for the F1-score to zero, indicating that the model collapses over all time steps. This was in the same vein as Tent \cite{tent} and CoTTA \cite{cotta} with this setup that showed the worst performance compared to the other baselines, as presented in Table~\ref{tab:comparison}.

Second, we modified the BN layers to track normalization statistics without initializing them. Subsequently, we normalized the input samples with updated statistics and transformed them with updated learnable parameters for prediction. Compared to the first configuration, this setup showed a performance improvement across all metrics. However, since the BN layers updated the normalization statistics based on the statistics of individual samples, the performance was still poor, which was consistent with the claim of \cite{sar}.

Lastly, we configured the BN layers with our proposed BN configuration, which used the fixed normalization statistics pretrained with the multiple source subjects. We normalized the input samples with fixed normalization statistics and transformed them with updated learnable parameters for prediction. This setup showed the best performance compared to the other configurations, which indicates that using fixed normalization statistics can enhance stability and adaptability when adapting to the target distribution for driver drowsiness detection systems.

\subsection{Configuration of the BN Layers for the TTA Baselines}
To demonstrate the effectiveness of the proposed BN configuration, we applied it to the existing TTA baselines that involve updating the BN layers \cite{tent, cotta, note, rotta, tta_MI}. Tent \cite{tent}, CoTTA \cite{cotta}, Wimpff \textit{et al.} \cite{tta_MI} normalized the input target data with estimated channel-wise statistics of them. On the other hand, NOTE \cite{note} and RoTTA \cite{rotta} updated the normalization statistics.

As shown in Fig.~\ref{fig:reconf}, the overall performance increases significantly after applying our proposed configuration for the BN layers. Specifically, Tent \cite{tent} and CoTTA \cite{cotta} not only significantly boost their average F1-scores but also surpass the average F1-score of the source baseline model. In addition, NOTE \cite{note} and Wimpff \textit{et al.} \cite{tta_MI} also show substantial improvements when applying our configuration. In contrast, RoTTA \cite{rotta} does not exhibit performance improvements, indicating that its removal criterion fails to filter out noisy and outdated samples effectively.

The significant performance improvements across different TTA methods highlight the adaptability and effectiveness of the proposed method. Overall, our findings indicate that the configuration of the BN layers is crucial for improving the performance and robustness to distribution shifts in EEG-based driver drowsiness detection framework.


\section{Discussion}
\subsection{Feature Representation Analysis of Adaptation Methods} 

In Fig.~\ref{fig:viz}, we present a UMAP visualization of features from target subjects (S5, S6, and S7) to compare different adaptation methods. Each row corresponds to a target subject, and each column represents a specific adaptation method.
The first column, showing the results of the empirical risk minimization (ERM) model without adaptation, reveals a significant overlap between the alert and drowsy states.

In the second and third columns, we present features adapted using existing BN configurations, labeled ``Initialize statistics'' and ``Track statistics''. Even after adapting the model to the target subjects, the boundaries between alert and drowsy states remain unclear, with substantial ambiguity between the classes.
The fourth and fifth columns present results from the TTA methods NOTE \cite{note} and Wimpff \textit{et al.} \cite{tta_MI}. These methods show moderate improvement in separating states; however, noticeable regions of overlap persist.
Finally, in the last column, we show the features from our proposed method, where the separation between the alert and drowsy states is significantly more distinct than in the other approaches. This clear distinction suggests that our method improves performance and feature disentanglement more effectively than existing methods, indicating its greater suitability for addressing the challenges of driver drowsiness detection.

\subsection{Limitation and Future Work}
The proposed framework demonstrates significant improvements in driver drowsiness detection within TTA environments. However, some limitations remain.
First, this study was conducted in a controlled simulator, which lacks real-world factors such as vehicle vibrations, changing weather, and external noise. Future work should validate the framework in real-world driving environments to ensure robustness against environmental noise and diverse sensory stimuli.
Second, we relied on binary labels derived from reaction time thresholds. Since drowsiness is inherently a continuous physiological phenomenon, this simplification may not fully capture transitional phases. Future studies could utilize fine-grained labels (e.g., Karolinska Sleepiness Scale \cite{dykim_ITS}) and explore regression or ordinal classification to better model the gradual progression of drowsiness.
Third, we employed a fixed 3-second time window for EEG segments. Future investigations could analyze the impact of varying window sizes to optimize the trade-off between detection latency and information content.
Finally, while this study focuses on drowsiness, our calibration-free framework is extensible to other EEG tasks \cite{buzzelli2023action, thomas2021midecoding, swlee_action}. Future research could explore its adaptability to broader BCI applications and wearable devices \cite{ronca2022validation, swlee_biometrics} in low-resource settings.

\section{Conclusion}
We proposed a novel driver drowsiness detection framework utilizing online TTA methods, addressing the variability of EEG signals.
We update only the learnable parameters of the BN layers with entropy minimization regularized by energy-bounded loss, while preserving the pretrained normalization statistics.
The memory bank helps maintain a dynamic view of the target domain by using a novel removal criterion based on the energy function and persistence time.
In addition, we integrate prototype learning for robust predictions against distribution shifts.
Extensive experiments validated the robustness of our proposed framework against the distribution shifts and demonstrated the effectiveness and reliability of online TTA methods for EEG-based driver drowsiness detection systems.

\bibliographystyle{IEEEtran} 
\bibliography{references}
\end{document}